\documentclass[graybox]{svmult}

\usepackage[hyphens]{url}

\usepackage{mathptmx}       
\usepackage{helvet}         
\usepackage{courier}        
\usepackage{type1cm}        
\usepackage{makeidx}         
\usepackage{graphicx}        
\usepackage{multicol}        
\usepackage[bottom]{footmisc}
\usepackage[numbers]{natbib}
\usepackage{amsmath}
\usepackage{bm}

\usepackage{subcaption}
\usepackage{float}
\usepackage{multirow}
\usepackage{wrapfig}
\usepackage{pgfplots}
\pgfplotsset{compat=1.9}
\usepackage{adjustbox}

\usepackage{pgfplots}
\usetikzlibrary{patterns}
\definecolor{orange}{RGB}{241,163,64}
\definecolor{white}{RGB}{247,247,247}
\definecolor{purple}{RGB}{153,142,195}

\usepackage[skip=2pt,font=small,labelfont=bf]{caption} 

\makeindex             

\begin{document}

\title*{Planning Multi-Fingered Grasps as Probabilistic Inference in a Learned Deep Network}
\author{Qingkai Lu, Kautilya Chenna, Balakumar Sundaralingam, and Tucker Hermans}
\authorrunning{Lu, Chenna, Sundaralingam, and Hermans}
\institute{Utah Robotics Center, School of Computing, University of Utah, Salt Lake City, UT, USA\\ \email{qklu,bala,thermans@cs.utah.edu; kautilya.chenna@utah.edu}}
%
%
\maketitle

%

\abstract{We propose a novel approach to multi-fingered grasp planning leveraging learned deep neural network models.
  We train a convolutional neural network to predict grasp success as a function of both visual information of an object and grasp configuration. We can then formulate grasp planning as inferring the grasp configuration which maximizes the probability of grasp success. We efficiently perform this inference using a gradient-ascent optimization inside the neural network using the backpropagation algorithm.
Our work is the first to directly plan high quality multi-fingered grasps in configuration space using a deep neural network without the need of an external planner. We validate our inference method performing both multi-finger and two-finger grasps on real robots. Our experimental results show that our planning method outperforms existing planning methods for neural networks; while offering several other benefits including being data-efficient in learning and fast enough to be deployed in real robotic applications.
}
\vspace{-24pt}
\section{Introduction and Motivation}
\vspace{-12pt}
\label{sec:intro}
Learning-based approaches to grasping~\cite{saxena2006learning,Saxena-aaai2008,lenz2015deep,pinto2016supersizing,Kopicki2016} have become a popular alternative to geometric~\cite{sahbani2012overview, bohg2014data, ciocarlie2007dexterous, dragiev2011gaussian} and model-based planning~\cite{Grupen1991,Murray1994} over the past decade. In particular grasp learning has shown to generalize well to previously unseen objects where only partial-view visual information is available. 
More recently, researchers have looked to capitalize on the success of deep neural networks to improve grasp learning. Broadly speaking deep neural network methods for grasp learning can be split into two approaches: predicting grasp success for an image patch associated with a gripper configuration~\cite{lenz2015deep,gualtieri2016high,pinto2016supersizing,levine2016learning,mahler2017dex,johns2016deep,varley2015generating} and directly predicting a grasp configuration from an image or image patch using regression~\cite{redmon2015real, kumra2016robotic, veres2017modeling}.
While these deep learning approaches have shown impressive performance for parallel jaw grippers (e.g.~\cite{pinto2016supersizing}) relatively little work has focused on the more difficult problem of multi-fingered grasping~\cite{varley2015generating,veres2017modeling,kappler2015leveraging}. We believe two primary difficulties restrict the use of deep learning for multi-fingered grasping (1) the input representation used for grasp configurations in neural networks and (2) the reliance on external planners for generating candidate grasps.

In order to combat these two problems, we propose an alternative approach to grasp planning with deep neural networks, where we directly use the learned network for planning. In our work, we train a network to predict grasp success; however, we use the trained network in a substantially different and novel way from currently employed sampling methods for grasp planning. The grasp representation we use fundamentally enables our approaches' success. 
In addition of giving the neural network the image patch, \(z\),  as input, we also provide the grasp configuration parameters, \(\theta\) (e.g. joint (preshape) angles, wrist pose, etc.).

Once trained, given a new object patch, \(z\), we perform inference over the grasp parameters, \(\theta\), in order to maximize the probability of grasp success \(P(Y=1|z,\theta)\) learned by our convolutional neural network (CNN). We perform this probabilistic inference as a direct optimization over \(\theta\) using constrained gradient-ascent, which leverages the efficient computation of gradients in neural networks, while ensuring joint angles remain within their limits.
Thus, our approach can quickly plan reliable multi-fingered grasps given an image of an object and an initial grasp configuration.

Our planner offers a number of benefits over previous deep-learning approaches to multi-fingered grasping. Kappler and colleagues~\cite{kappler2015leveraging} learn to predict if a given palm pose will be successful for multi-fingered grasps using a fixed preshape and perform planning by evaluating a number of sampled grasp poses. Varley et al.~\cite{varley2015generating} present a deep learning approach to effectively predict a grasp quality metric for multi-fingered grasps, but rely on an external grasp planner to provide candidate grasps.
In contrast, our method learns to predict grasp success as a function of both the palm location and preshape configuration and plans grasps directly using the learned network. Saxena et al.~\cite{Saxena-aaai2008} also perform grasp planning as inference using learned probabilistic models; however they use separate classifiers for both the image and range data, using hand selected models instead of a unified deep model. Zhou and Hauser~\cite{zhou6dof} concurrently propose a similar optimization-based grasp planning approach to ours using a similar CNN architecture. In contrast to our work, they do not interpret planning as probabilistic inference; they optimize only for hand pose, ignoring hand joint configurations; and they validate only in simulation.

Veres et al.~\cite{veres2017modeling} train a conditional variational auto-encoder (CVAE) deep network to predict the contact locations and normals for a multi-fingered grasp given an RGBD image of an object. In order to perform grasping an external inverse kinematics solver must be used for the hand to try and reach the desired contact poses as best as possible. Implicit in such a regression method as proposed in~\cite{veres2017modeling} lies the assumption that there exists a unique best grasp for a given object view.
In contrast, our method can plan multiple high quality grasps for a given object using different initial configurations. This offers the robot the option of selecting a grasp best suited for its current task. Additionally, we show that our classification-based network can effectively learn with a smaller dataset compared with a regression network, which can not leverage negative grasp examples.

We propose two novel CNN architectures to encode grasp configurations for our planner, but many alternatives would surely also work. Our proposed multi-channel network (c.f.~\ref{sec:config-cnn}) has a similar architecture to that of~\cite{levine2016learning}, but we train on joint configurations instead of motor commands and use only a single image as input, as we evaluate grasps and do not learn a controller to perform grasping. Our alternative patch-based network described in Section~\ref{sec:patch-cnn} was inspired by the fingertip patches used in~\cite{varley2015generating}; however, we construct our patches as a function of the grasp parameters differently in order to improve learning and efficiently perform gradient ascent.



The contributions of this paper are as follows:
\begin{itemize}\vspace{-6pt}
\item{We present a method for performing grasp planning as probabilistic inference in a deep neural network, that}
  \vspace{-6pt}
\begin{itemize}
\item{is the first work to directly optimize over grasp configurations inside a learned deep network;}
\item{is data-efficient compared with direct regression CNNs;}
\item{can naturally predict a variable number of high quality grasps;}
\item{can improve initial grasp configurations which would fail to result in successful grasps;}
\item{requires far fewer grasps as initializations than sampling-based approaches.}
    \vspace{-6pt}
\end{itemize}
\item{We propose two novel CNN architectures for use with our planning algorithm.}
\item{We provide a multi-finger grasp dataset from simulation with more realistic data than currently existing simulation datasets.}
\item{We experimentally validate the effectiveness and efficiency of our inference method for both multi-finger and two-finger grasp planning on real robots.}
 \vspace{-6pt}
\end{itemize}

In the next section we provide a formal description of our grasp planning approach. We follow this in Section~\ref{sec:grasp_learning} with an overview of our approach to multi-finger grasp learning and the novel CNN architectures for predicting grasp success. We then give a thorough account of our experiments and results in Section~\ref{sec:exp}. We conclude with a brief discussion in Section~\ref{sec:conclusions}.


\vspace{-24pt}
\section{Grasp Planning as Probabilistic Inference}
\vspace{-12pt}
\label{sec:grasp_inference}
Following~\cite{ciocarlie2007dexterous} we define the grasp planning problem as finding a grasp pre-shape configuration~(hand joint angles and palm pose). The robot then moves to this pose and executes a controller to close the hand forming the grasp on the object. We focus on scenarios where a single object of interest in isolation exists in the scene. Importantly, we assume no explicit knowledge of the object beyond this sensor reading.
The problem we address states, given such a grasp scenario, plan a grasp preshape configuration that allows the robot to successfully grasp and lift the object without dropping it.


We propose planning grasps by finding the grasp configuration which maximizes the grasp success probability. We use a deep neural network to predict the probability of grasp success, \(Y\), as a function of the sensor readings, \(z\), and hand configuration, \(\theta\). We formalize probabilistic grasp inference as an optimization problem:
\begin{equation}
\begin{aligned}
& \underset{\theta}{\text{argmax}}
& & p(Y=1 | \theta, z, w) = f(\theta, z, w) \\
& \text{subject to}
& & {\theta}_{min} \leq \theta \leq {\theta}_{max}.
\end{aligned}
\label{eq:inf_obj}
\end{equation}
In Eq.~\ref{eq:inf_obj} \(f(\theta, z, w)\) defines a neural network classifier with logistic output trained to predict the grasp success probability as a Bernoulli distribution over \(Y\). The parameters $w$ define the neural network parameters. We encode joint limits and the reachable workspace of the robot hand as constraints on the decision variables. We can now directly optimize over \(\theta\) in order to infer the maximum likelihood grasp estimate, \(\theta^{\star}\).

In order to efficiently solve this optimization we use a gradient-ascent approach, which leverages the structure of the neural network to efficiently compute gradients. The gradients are computed using the well-known backpropogation algorithm with respect to the grasp input, $\theta$, instead of the more typical optimization of network parameters, \(w\), during learning. The complete algorithm takes as input the current object image, \(z\), and an initial grasp configuration, \(\theta^0\). We compute the success probability by evaluating a forward pass of the neural network, then update the grasp parameters, \(\theta\), using backpropogation in backwards pass through the network. We iterate these forward and backward passes until convergence.
We handle the linear constraints using gradient projection and apply a backtracking line search to determine the gradient step length at each iteration.
Initial hand configurations could be generated in a number of different ways, we describe the heuristic approach we use in Section~\ref{subsec:grasp_system}.

Our formulation allows for a number of straightforward extensions, which we do not consider in this work. First, adding a prior over \(\theta\) would allow one to infer the Maximum a posteriori (MAP) grasp estimate, which could also be learned from data. This could encode, for example, generally effective preshapes independent of the object. Additionally other constraints could be added to the optimization such as collision avoidance constraints or the full forward kinematics of the robot arm. Finally, other sensing modalities could be given as input to the neural network if available such as tactile or haptic information.
In the next section we define  two neural network architectures we use in evaluating our approach. However, many other networks would likely work well within our framework, as long as they use the same inputs and outputs required by our planner.


\vspace{-24pt}
\section{Deep Networks for Multi-fingered Grasp Learning}
\vspace{-12pt}
\label{sec:grasp_learning}
We present two novel neural network architectures designed to predict the probability of grasp success for multi-fingered hands. Importantly, the networks operate as a function of both a grasp configuration and RGBD image of the object of interest.
For a given RGBD image we compute the surface normals and curvature from the associated point cloud giving an 8-channel representation (i.e. RGB (3), depth (1), normal (3), and curvature (1)).
In the remainder of this section we first detail the specifics of the two networks. We then explain the training algorithm used, followed by a brief discussion of the gradient computation necessary to perform the inference described in the previous section.


\vspace{-24pt}
\subsection{A Multi-channel Grasp Configuration and Image CNN}
\label{sec:config-cnn}
\begin{figure}
  \vspace{-3mm}
  \includegraphics[width=1.0\linewidth, height=0.5\linewidth]	{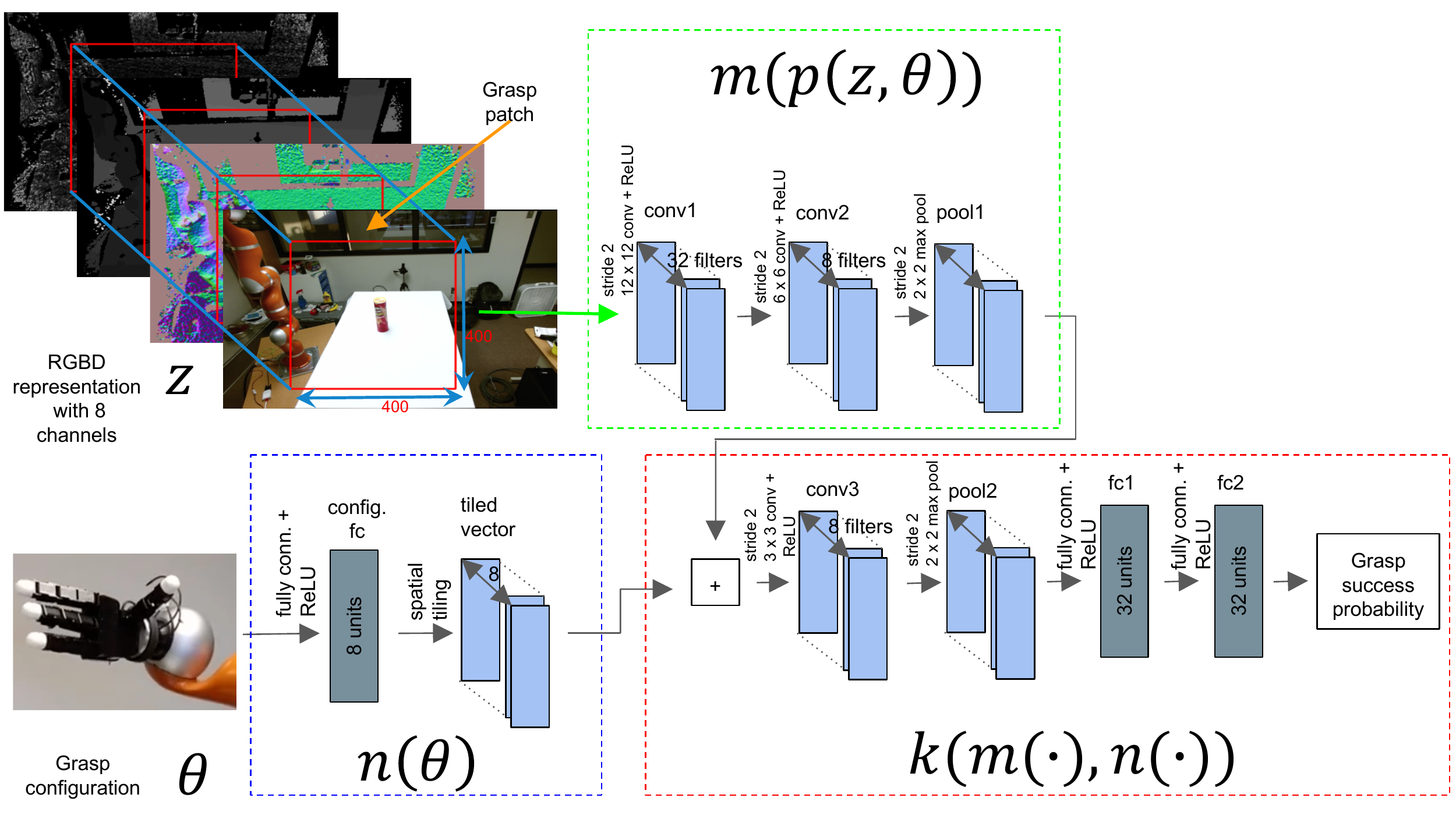}
  \caption{The config-CNN architecture. Top left: the RGBD representation, \(z\), extracted grasp patch, \(p\), and image patch network channel \(m(\cdot)\). Bottom: the Allegro hand and grasp configuration network channel \(n(\cdot)\). Right: the combined feature network layers and output \(k(\cdot)\).}
  \label{fig:config_cnn}
  \vspace{-12pt}
\end{figure}

\vspace{-12pt}
We define a multi-channel deep neural network which takes as input a grasp configuration and RGBD image grasp patch and predicts as output the probability of grasp success. We extract a $400\times400$ pixel object patch from the 8-channel RGBD representation. In our experiments we examine both keeping this object patch fixed and moving it as a function of the palm pose of the grasp.
The configuration RGBD CNN takes the grasp patch and the grasp preshape configuration as the input to learn to predict the grasp success probability.
We pass the image patch through two convolution layers and one max pooling layer.
We first process the grasp configuration, $\theta$, through one fully connected layer. We tile the resulting grasp configuration features point-wise across the spatial dimensions of the response map of the \emph{pool1} features output from the image feature channel (c.f.~\cite{levine2016learning}). This generates a concatenation of the grasp configuration and image features which pass through one convolution layer, then one max pooling layer, followed by two-fully connected layers, and a final logistic regression output layer. Figure~\ref{fig:config_cnn} shows this ``config-CNN'' structure in detail. 
The config-CNN contains 61k parameters in total.



\vspace{-36pt}
\subsection{An Image Patch Based CNN for Grasping}
\label{sec:patch-cnn}
\begin{figure}
  \vspace{-3mm}
  \centering
  \includegraphics[width=\linewidth]
  {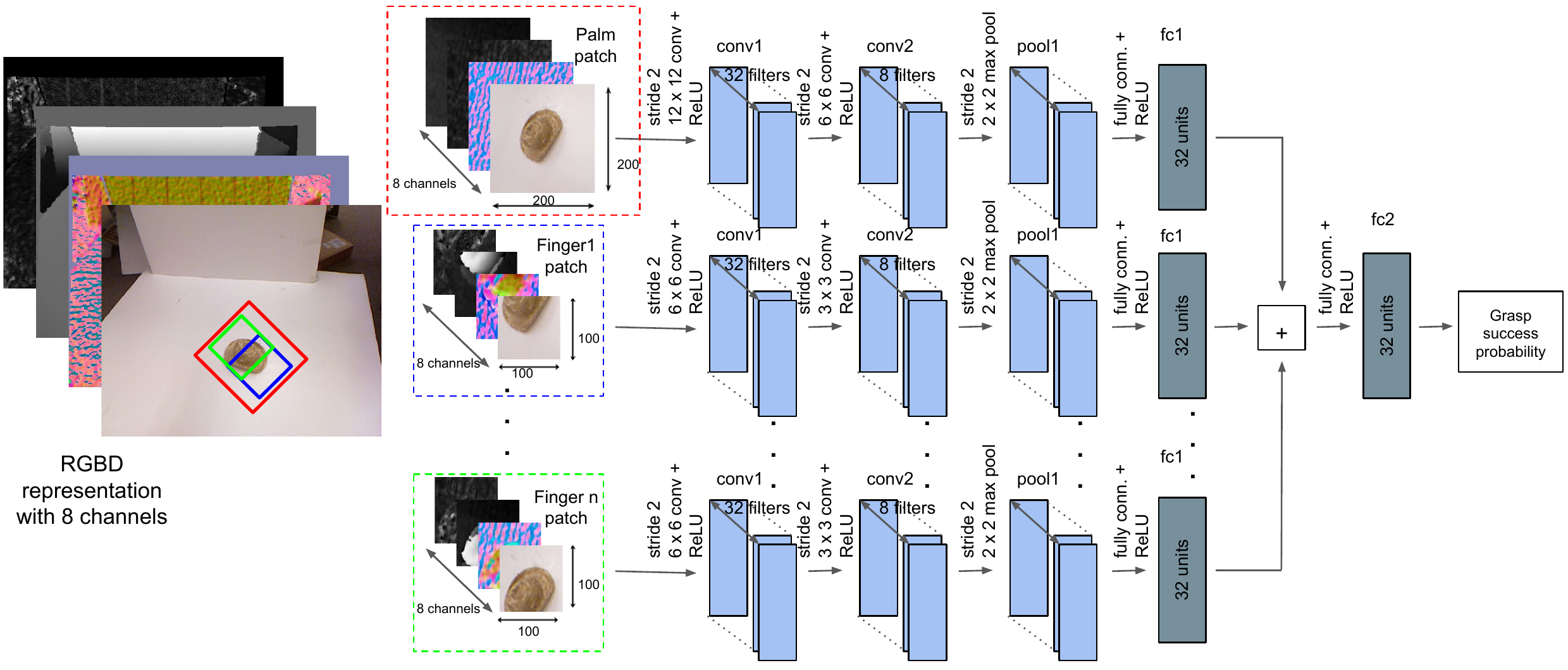}
  \caption{The patches-CNN architecture and patch input. Finger and palm patches are extracted from the 8-channel RGBD representation on the left. The red bounding box represents the palm patch. The blue and green boxes define the patches for the two Baxter gripper fingers.}
  \label{fig:patches_cnn}
\end{figure}

\vspace{-12pt}
As an alternative to the config-CNN, we define an image based network inspired by~\cite{varley2015generating}. We extract image patches related to the palm pose and fingertip locations for a given grasp configuration, \(\theta\). Each image patch is passed through two convolution layers, one max pooling layer, and one fully connected layer~\emph{fc1}. We concatenate the features from each finger and palm~\emph{fc1} and feed these into one fully connected layer followed by  a logistic regression output.  We show the structure of the ``patches-CNN'' in Figure~\ref{fig:patches_cnn} implemented for the Baxter parallel jaw gripper as an example. However, extending it for use on a multi-finger hand is straightforward.

We use the same 8 channel image representation as with the config-CNN. We extract a $200\times200$ pixel patch centered at the projected palm location and oriented to align with the project palm orientation. We extract $100\times100$ pixel patches for each finger centered at the projected fingertip location in the image rotated to align with the projected palm orientation. 
Our patches-CNN has 316k parameters in total.


\vspace{-12pt}
\subsection{Training}
\vspace{-12pt}
We train our classifiers using the common cross entropy loss function:
\begin{equation}
\begin{aligned}
& \underset{w}{\text{argmin}}
& & \sum_{i=1}^{M} -y_i \log(f(\theta, z, w)) - (1 - y_i) \log(1 - f(\theta, z, w))  \\
\end{aligned}
\label{eq:learn_objective}
\end{equation}
where \(y_i\) defines the ground truth grasp label as 1 if successful and 0 if failed. We optimize for training by stochastic gradient descent (SGD) using Adam optimizer with mini-batches of size $8$ for $6,000$ iterations. The learning rate starts at $0.001$ and decreases by $10\times$ every $2,000$ iterations. We apply dropout to prevent overfitting keeping weights with a probability of $0.75$ and apply Xavier initialization~\cite{glorot2010} for the weights. For the multi-finger grasp CNN training, we oversample positive grasps making sure at least one positive grasp exist in each mini-batch. The same training set up is used for the patches-CNN, except we train for $60,000$ iterations with the learning rate decreasing by $10$x every $20,000$ iterations. When training on the two-fingered grasps dataset from~\cite{lenz2015deep} we do not oversample, but mirror each grasp patch to double the number of training examples. We implemented and trained all of our neural network models using TensorFlow (\url{http://tensorflow.org/.}).
\vspace{-24pt}
\subsection{Computing Gradients for Image Patches in Inference}
\vspace{-12pt}
Since, we define the patches used as input to our CNNs, as a function of the grasp configuration, \(\theta\), we must compute the gradient of the patch with respect to the configuration parameters for use in inference.
In the patches-CNN the gradient takes the form:
\begin{equation}
\frac{\partial f}{\partial \theta} = \sum_{i=0}^{N} \frac{\partial f}{\partial p_i}\frac{\partial p_i}{\partial \theta}
\label{eq:patches_cnn_gradient}
\end{equation}
where we define the sum over the derivatives with respect to the image patches $p_i$ associated with the \(N-1\) fingers and palm.

We can decompose the config-CNN \(f(\cdot)\)into sub-modules \(m\), \(k\), and \(n\) so that $f=k(m(p(z, \theta)),n(\theta))$ as shown in Fig.~\ref{fig:config_cnn}. This gives the following equation in computing the gradient:
\begin{equation}
\frac{\partial f}{\partial \theta} = \frac{\partial k}{\partial n}\frac{\partial n}{\partial \theta} + \frac{\partial k}{\partial m}\frac{\partial m}{\partial p}\frac{\partial p}{\partial \theta}
\label{eq:config_cnn_gradient}
\end{equation}

We use backpropogation to compute most of the terms in Equations~\ref{eq:patches_cnn_gradient} and~\ref{eq:config_cnn_gradient}; however, for both networks we use finite differences to estimate the patch gradient with respect to the hand configuration:
\begin{equation}
\frac{\partial p}{\partial \theta} = \frac{p(z, \theta + \epsilon) - p(z, \theta - \epsilon)}{2\epsilon}
\label{eq:img_conv_gradient}
\end{equation}

%



\section{Experimental Validation}
\vspace{-12pt}
\label{sec:exp}
In this section, we describe our simulation-based training data collection process and the experimental evaluation of our grasp inference approach. We conduct multi-finger grasp experiments using the same four-fingered Allegro hand mounted on a Kuka LBR4 arm in simulation and on the real robot. We compare to a heuristic grasp procedure, as well as the two dominant approaches to deep-learning based grasp planning: sampling and regression. Finally, we show the applicability of our inference method to two-finger grippers by preforming real-world experiments on the Baxter robot. All data and software used in this paper are available for use at: \url{https://robot-learning.cs.utah.edu/project/grasp_inference}.

\vspace{-24pt}
\subsection{Multi-finger Grasping Data Collection}
\label{subsec:grasp_system}
\begin{wrapfigure}{r}{0.3\textwidth}
  \vspace{-18pt}
    \centering
    \begin{subfigure}[b]{0.3\textwidth}
        \includegraphics[width=\textwidth]{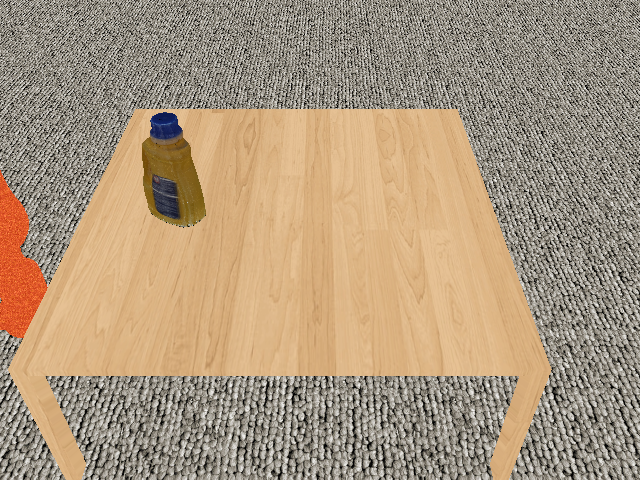}
        \caption{The RGB image.}
        \label{fig:rgb_blensor}
    \end{subfigure}
    \begin{subfigure}[b]{0.3\textwidth}
        \includegraphics[width=\textwidth]{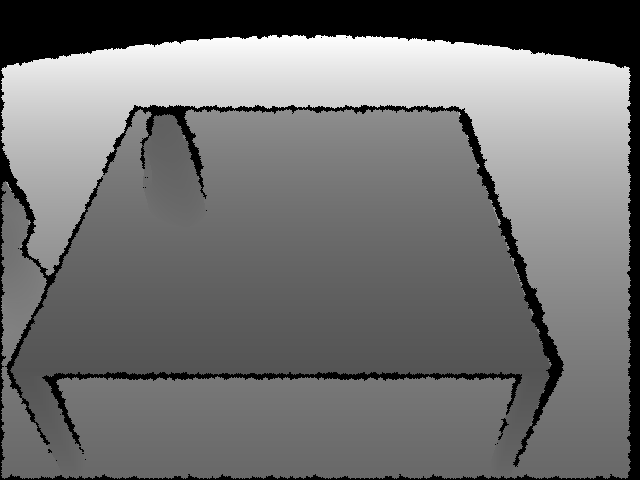}
        \caption{The depth image.}
        \label{fig:depth_blensor}
    \end{subfigure}
    \caption{The RGB and depth image generated by Blensor for one grasp trial of the ``detergent'' object. The bottom part of the orange robot arm can be seen in the left side of the image.}
    \label{fig:image_blensor}
\end{wrapfigure}

\vspace{-12pt}
We developed a grasping system using ROS that runs both in simulation and on the real robot. We collected simulated grasps data using the Allegro hand mounted on the Kuka LBR4 arm inside the Gazebo simulator with the DART physics engine (https://dartsim.github.io/). We use Blensor~\cite{gschwandtner2011blensor} to generate RGB and depth images simulating a Kinect camera we use in real-world experiments. Example images generated by blensor can be seen in Figure~\ref{fig:image_blensor}.

For a given trial we place the selected object, with a predetermined support surface facing down, at a location chosen uniformly at random from a \(0.2\times0.2\)m rectangle area with a uniformly random orientation.
We then perform object segmentation by fitting a plane to the table using RANSAC and extract the points above the table. We compute a 3D bounding box of the segmented object using PCA used to generate heuristic grasps associated with the bounding box's top face and its two faces closer to the camera.

We generate a grasp for a given face by setting the palm to be a fixed distance from the face center (2cm for top grasps and 3cm for side grasps) in the direction of its surface normal. For data collection we add zero mean Gaussian noise with a standard deviation of 1mm along the normal to get a more diverse set of grasps.
We rotate the palm to align with the bounding box face. For side grasps we have the thumb pointing towards the top, for top grasps we randomly select to align with either the major or minor axis of the face and add an additional small amount of random noise.
We generate a preshape by randomly sampling joint angles for the first two joints of all fingers within a reasonable range, fixing the last two joints of each finger to be zero. The preshape of the Allegro hand has 14 parameters: 6 for the palm pose and 8 for joint angles. In data collection, each preshape reachable for the arm using the RRT-connect motion planner in  Moveit is executed and recorded. In experiments, all 3 preshapes are used as initializations for inference.
For grasping the object we first use a position controller to move the finger preshape joints to the desired preshape, followed by moving the arm to reach the desired palm preshape pose. We then use a simple grasp controller to close the hand. The controller closes the fingers at a constant velocity stopping each finger independently when contact is detected by the measured joints velocities being close to zero.
We found slightly different controllers to work well for top grasps versus side grasps.
For most grasps the controller closes the last three joints of non-thumb fingers and the last two joints of the thumb. Importantly, the last three joints includes both preshape and non-preshape joints. However, for overhead grasps the controller closes only the second joint of non-thumb fingers and only the third joint of the thumb.

If after closing the hand the robot can lift the object  to a height 0.15m without dropping it, the system labels the grasp as successful.
We used this grasp system to collect a dataset containing 1507 grasp trials. The dataset covers all 125 objects of the Bigbird~\cite{singh2014bigbird} dataset with an average of 12 grasps per object, depending on how many grasps were reachable by the planner.
Of the collected grasps attempted 159 were successful.
\vspace{-14pt}
\subsection{Grasp Classification CNN Evaluation}
\vspace{-12pt}
We first compare the performance of our two proposed network architectures config-CNN and patches-CNN. We train the networks using a random 80\% of the collected data and test with the remaining 20\%. We repeat this procedure to perform a five-fold cross validation. 
We compare two different cross validation scenarios. In the first the training set contains examples of all objects in the test set. In the second scenario objects in the testing set are held out from the training set to simulate observing novel objects. We refer to these scenarios as ``seen'' and ``unseen'' respectively. In both scenarios, each fold contains approximately the same percentage of successful grasps as the complete data-set.

The patches-CNN performs poorly for the classification of multi-finger grasps for both seen and unseen scenarios. The network predicts all test grasps as failure cases. Different training parameters (e.g. learning rate, dropout probability, the number of batches, the number of training iterations, and the oversampling of successful grasps) were attempted to improve learning, including using the successful settings from the config-CNN.  As such we do not report detailed results from the patches-CNN.
The comparative success of the config-CNN shows that having direct access to the grasp configuration parameters allows the network to learn a more useful feature representation than providing this information indirectly through the image patch.
\begin{wrapfigure}{r}{0.6\textwidth}
  \vspace{-24pt}
    \centering
	\includegraphics[width=0.6\textwidth]{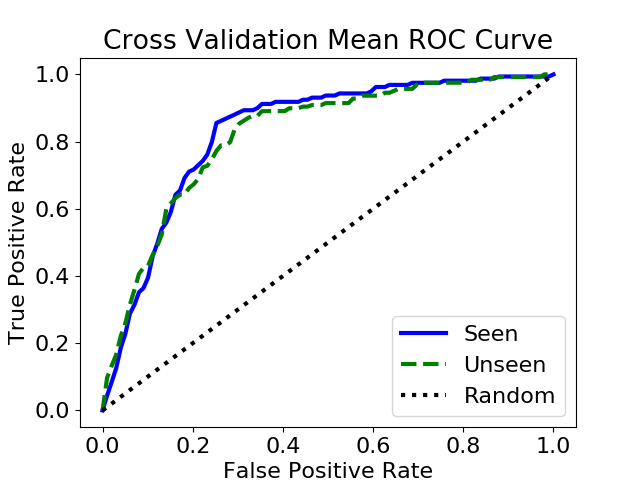}
	\caption{The average ROC curve of five-fold cross validation for seen and unseen objects. For seen objects, the AUC mean is: 0.83 the AUC std: 0.02.  For unseen objects, the AUC mean is: 0.82, the AUC std: 0.06.  For random guessing, the AUC mean is: 0.5.}
	\label{fig:roc_seen_unseen}
 \vspace{-24pt}
\end{wrapfigure}
The average receive operator characteristic (ROC) curve and the area under the ROC curve (AUC) for the config-CNN can be seen in Figure~\ref{fig:roc_seen_unseen} along with the performance of selecting a label at 50\% probability for comparison.
Table~\ref{tab:mectrics_cv} shows the mean and standard deviation of the accuracy and F1 score across all validation folds for both seen and unseen objects.
We treat predictions with grasp probability above 0.4 as positive, based on the ROC curve and our desire for high recall.
As we can see, the config-CNN predicts multi-finger grasp success reasonably well for both seen and unseen objects, significantly outperforming chance.

\vspace{-2mm}

\begin{table}
  \vspace{-2mm}
\begin{center}
  \caption{Accuracy and F1 score of the RGBD config-CNN of the 5-fold cross validation for seen and unseen objects. We list the mean with standard deviation in parentheses.}
\label{tab:mectrics_cv}
\begin{tabular}{p{2cm}p{3cm}p{3cm}}
\hline\noalign{\smallskip}
Experiment & Accuracy & F1 \\
\noalign{\smallskip}\noalign{\smallskip}
  Seen & 0.766  (0.038) & 0.405 (0.016) \\
  Unseen & 0.756  (0.05) & 0.385  (0.096) \\
  Random & 0.5 (-)  & 0.172  (0.02) \\
\noalign{\smallskip}\hline\noalign{\smallskip}
\end{tabular}\\
\end{center}
  \vspace{-5mm}
\end{table}
\vspace{-3mm}

As an additional comparison to our proposed networks we implemented a regression network to directly predict the grasp configuration parameters, \(\theta\), as a function of the 8-channel RGBD object patch. We implemented a network architecture similar to the RGBD channel of our config-CNN, except that we removed the third convolution layer used for concatenating the configuration and grasp patch features. The output layer of the regression CNN consists of separate linear regression outputs for each grasp parameter. We apply ridge regression to regularize the network with a strength of 0.5 in addition to the standard least-squares loss function.

In order to predict good grasps, the regression CNN can only learn from the 159 successful grasp examples in our dataset and makes no use of the remaining 1348 samples. Unsurprisingly this is far too few samples to reasonably learn in our regression CNN with 90k parameters. The regression CNNs in~\cite{redmon2015real, kumra2016robotic} have at least 60M parameters and the CVAE deep learning model in~\cite{veres2017modeling} has more than 2M parameters, thus we find it highly unlikely that they will work with our data-set.

We further validate the regressors performance by examining the distance between the predicted grasp and the ground truth example for a given object pose.
The mean Euclidean distance between the predicted preshape palm location and the successful ground truth preshape palm location was above 0.4m and the minimum distance was above 0.2m. Thus the grasp regression CNN cannot predict palm locations within an effective distance of the object for grasping. The predicted palm orientations and joint angles were also far from the ground truth preshapes. We examined alternative deeper and shallower network structures for regression, but found none which could effectively learn from such a limited amount of data.




\vspace{-24pt}
\subsection{Multi-finger Grasp Inference in Simulation}
\vspace{-12pt}
We first evaluate our inference procedure for multi-finger grasp planning by performing grasping experiments on 10 objects of daily life from the Bigbird dataset.
We test grasps for each object at 5 different random poses on the table. For each pose we generate three initial grasp configurations following the heuristic described in Section~\ref{subsec:grasp_system}. We apply our CNN gradient ascent planner to each initialization and select the grasp preshape configuration with highest predicted grasp success probability for execution on the robot. 
The robot automatically selects to close with the top or side grasp controller depending on which bounding box face was used for initialization.
If the planned grasp is not reachable by the arm motion planner, then we generate a new random object pose for evaluation.

We used the following parameters in performing inference.
We constrained the palm location to be within a \(0.1\times 0.1\times 0.1\)m cube around the palm location given as initialization. We restrict the palm orientation to stay within 0.3 rad of the initial pose expressed in XYZ Euler angles. All preshape joint angles were restricted to the same limits used with the heuristic grasp. The maximum number of iterations for the inference is 100. We set the initial step size for backtracking line search to 0.001 and allow it a maximum of 10 iterations for each gradient ascent step. We set the control parameter of backtracking line search to 0.5.

We found that changing the object image patch as a function of the palm pose did not significantly change the grasps resulting from inference. As such, we only compute the change in grasp configuration using the gradient $\frac{\partial k}{\partial n}\frac{\partial n}{\partial \theta}$ (c.f. Sec.~\ref{sec:config-cnn}). This approximation sped up the inference procedure significantly without noticeably changing the grasps generated by the planner.
\begin{figure}
  \vspace{-3mm}
  \begin{adjustbox}{minipage={\textwidth}}
    \begin{tikzpicture}
      \begin{axis}[
        ybar, ymin=-1, ymax=100,
        ylabel={Seen Success Rate (\%)},
        axis lines*=left,
        symbolic x coords={Coffee Mate, Pringles, Syrup, Detergent, Canon Box, Body wash, Hot sauce, Mom to mom, Volumizing Shampoo, Spray Adhesive, All},
        xtick=\empty,
        ticklabel style = {font=\scriptsize, rotate=45},
        legend style={font=\scriptsize, draw=none, fill=none},
        ylabel style = {font=\scriptsize},
        bar width = 3.5pt, height=4.5 cm, width=\linewidth,
        legend style={area legend, anchor=north east, legend columns=4, },
        legend image code/.code={%
         \draw[#1] (0cm,-0.1cm) rectangle (2mm,1mm);},
        ]
        \addplot [fill=orange, postaction={pattern=north east lines}] coordinates {
          (Coffee Mate,27)
          (Pringles,43)
          (Syrup,40)
          (Detergent,9)
          (Canon Box,20)
          (Body wash,53)
          (Hot sauce,17)
          (Mom to mom,42)
          (Volumizing Shampoo,27)
          (Spray Adhesive,27)
          (All,31)};

        \addplot [fill=yellow, postaction={pattern=horizontal lines}] coordinates {(Coffee Mate,60) (Pringles,60) (Syrup,40) (Detergent,0) (Canon Box,80) (Body wash,80) (Hot sauce,20) (Mom to mom,60) (Volumizing Shampoo,60) (Spray Adhesive,20) (All,48)};

        \addplot [fill=red, postaction={pattern=crosshatch}] coordinates {(Coffee Mate,20) (Pringles,40) (Syrup,60) (Detergent,0) (Canon Box,40) (Body wash,20) (Hot sauce,40) (Mom to mom,0) (Volumizing Shampoo,20) (Spray Adhesive,40) (All,28)};

        \addplot [fill=purple] coordinates {(Coffee Mate,80) (Pringles,80) (Syrup,60) (Detergent,20) (Canon Box,40) (Body wash,80) (Hot sauce,40) (Mom to mom,60) (Volumizing Shampoo,80) (Spray Adhesive,80) (All,62)};
        \legend{Heuristic, Max-eval, Sampling, Inference}
      \end{axis}
    \end{tikzpicture}
  \end{adjustbox}
  \begin{adjustbox}{minipage={\textwidth}}
    \begin{tikzpicture}
      \begin{axis}[
        ybar, ymin=-1, ymax=100,
        ylabel={Unseen Success Rate (\%)},
        axis lines*=left,
        symbolic x coords={Coffee Mate, Pringles, Syrup, Detergent, Canon Box, Body wash, Hot sauce, Mom to mom, Volumizing Shampoo, Spray Adhesive, All},
        xtick=data,
        ticklabel style = {font=\scriptsize, rotate=45},
        legend style={font=\scriptsize, draw=none, fill=none},
        ylabel style = {font=\scriptsize},
        bar width = 3.5pt, height=4.5 cm, width=\linewidth,
        legend style={area legend, anchor=northeast, legend columns=4, },
        legend image code/.code={%
         \draw[#1] (0cm,-0.1cm) rectangle (2mm,1mm);},
        ]
        \addplot [fill=orange, postaction={pattern=north east lines}] coordinates {
          (Coffee Mate,33) (Pringles,57) (Syrup,63) (Detergent,20) (Canon Box,0) (Body wash,27) (Hot sauce,23) (Mom to mom,44) (Volumizing Shampoo,33) (Spray Adhesive,42) (All,34)};

        \addplot [fill=yellow, postaction={pattern=horizontal lines}] coordinates {
          (Coffee Mate,40) (Pringles,60) (Syrup,80) (Detergent,0) (Canon Box,20) (Body wash,60) (Hot sauce,20) (Mom to mom,60) (Volumizing Shampoo,20) (Spray Adhesive,60) (All,42)};

        \addplot [fill=red, postaction={pattern=crosshatch}] coordinates {(Coffee Mate,0) (Pringles,0) (Syrup,40) (Detergent,0) (Canon Box,0) (Body wash,60) (Hot sauce,20) (Mom to mom,40) (Volumizing Shampoo,0) (Spray Adhesive,20) (All,18)};

        \addplot [fill=purple] coordinates {(Coffee Mate,40) (Pringles,80) (Syrup,40) (Detergent,40) (Canon Box,20) (Body wash,100) (Hot sauce,20) (Mom to mom,60) (Volumizing Shampoo,40) (Spray Adhesive,80) (All,52)};
      \end{axis}
    \end{tikzpicture}
  \end{adjustbox}
  \caption{\textbf{Simulation Success Rate Results:} Success rate for executed grasps on previously seen objects (top) and previously unseen objects (bottom) in simulation.\label{fig:sim_results_seen}}
\end{figure}
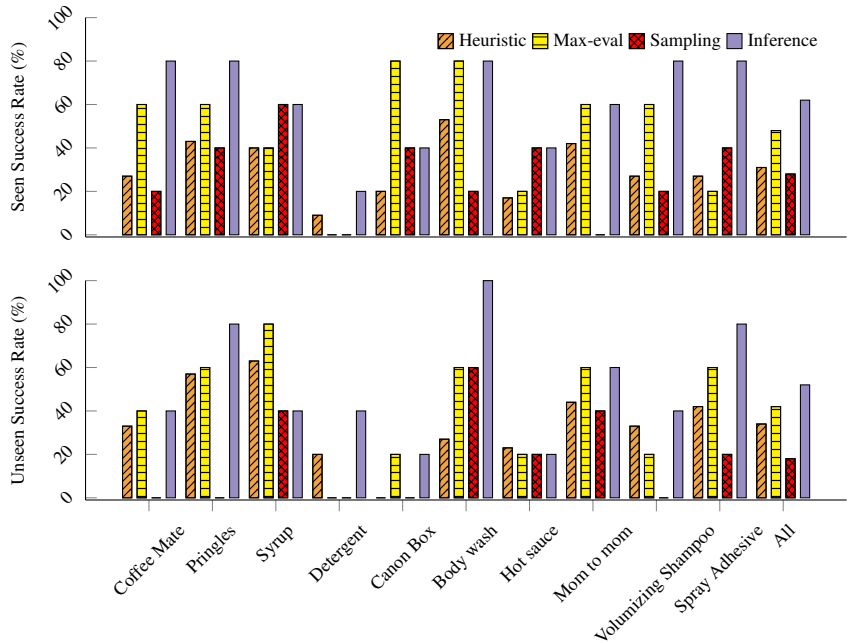

We examine our grasp planner on both seen and unseen objects. For the seen scenario we train the learner on the entire training set. In the unseen case we only hold out the test object from the training set to simulate the robot viewing a novel object.
We compare our planner to three baselines. First we perform the three heuristic grasps for each object pose used by the planner. We ignore cases where the arm planner could not find a plan to reach the grasp pose. This gives a maximum of 15 heuristic grasps per object.
The second approach evaluates the three heuristic grasps using the learned network and selects the one with highest predicted success probability to execute. We term this approach ``max-eval'' and executed it on an additional 5 random object poses per object.
Our final comparison performs a sampling approach where we generate 150 random grasp configurations. We evaluate all 150 generated grasps with our learned network and execute the grasp with highest predicted probability. Unlike the other three methods, the heuristic approach produced a variable number of reachable grasps for each object pose. In total we collected 121 heuristic grasp attempts for seen and 110 for unseen. 

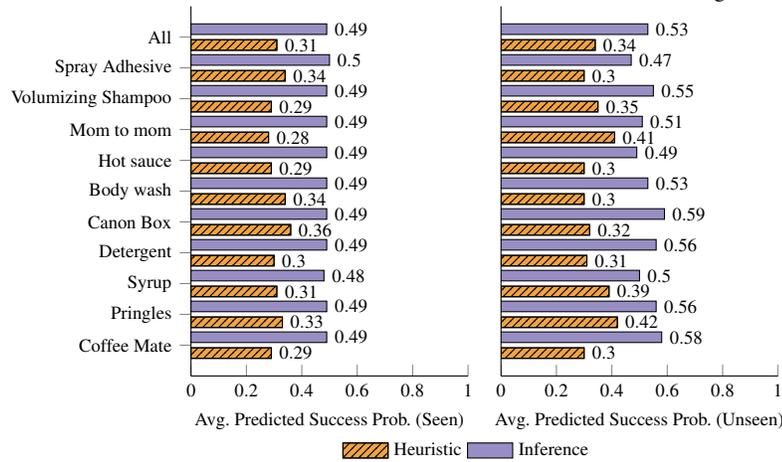
\begin{figure}[h!]
  \vspace{-3mm}
  \begin{adjustbox}{valign=t,minipage={.45\textwidth}}
    \begin{tikzpicture}
      \begin{axis}[
        xbar, xmin=0, xmax=1,
        xlabel={Avg. Predicted Success Prob. (Seen)},
        axis lines*=left,
        symbolic y coords={Coffee Mate, Pringles, Syrup, Detergent, Canon Box, Body wash, Hot sauce, Mom to mom, Volumizing Shampoo, Spray Adhesive, All},
        ytick=data,
        ticklabel style = {font=\scriptsize, /pgf/number format/fixed},
        legend style={font=\scriptsize, draw=none, fill=none},
        xlabel style = {font=\scriptsize},
        bar width = 4pt, height=6.5 cm, width=\linewidth,
        legend style={at={(1.0,-0.15)}, area legend,
          anchor=north,legend columns=-1},
        legend image code/.code={%
          \draw[#1] (0cm,-0.1cm) rectangle (0.6cm,0.1cm);},
        nodes near coords,
        nodes near coords align={horizontal},
        every node near coord/.append style={font=\scriptsize, /pgf/number format/fixed, /pgf/number format/precision=2},
        ]
        \addplot [fill=orange, postaction={pattern=north east lines}] coordinates {(0.29,Coffee Mate) (0.33,Pringles) (0.31,Syrup) (0.3,Detergent) (0.36,Canon Box) (0.34,Body wash) (0.29,Hot sauce) (0.28,Mom to mom) (0.29,Volumizing Shampoo) (0.34,Spray Adhesive) (0.31,All)};

        \addplot [fill=purple] coordinates {(0.49,Coffee Mate) (0.49,Pringles) (0.48,Syrup) (0.49,Detergent) (0.49,Canon Box) (0.49,Body wash) (0.49,Hot sauce) (0.49,Mom to mom) (0.49,Volumizing Shampoo) (0.5,Spray Adhesive) (0.49,All)};
        \legend{Heuristic, Inference}
      \end{axis}
    \end{tikzpicture}
  \end{adjustbox}
  \hfill
  \begin{adjustbox}{valign=t,minipage={.45\textwidth}}
    \begin{tikzpicture}
      \begin{axis}[
        xbar, xmin=0, xmax=1,
        xlabel={Avg. Predicted Success Prob. (Unseen)},
        axis lines*=left,
        symbolic y coords={Coffee Mate, Pringles, Syrup, Detergent, Canon Box, Body wash, Hot sauce, Mom to mom, Volumizing Shampoo, Spray Adhesive, All},
        ytick=\empty,
        ticklabel style = {font=\scriptsize, /pgf/number format/fixed},
        legend style={font=\scriptsize, draw=none, fill=none},
        xlabel style = {font=\scriptsize},
        bar width = 4pt, height=6.5 cm, width=\linewidth,
        legend style={at={(0.5,1.0)}, area legend,
          anchor=south,legend columns=-1},
        legend image code/.code={%
          \draw[#1] (0cm,-0.1cm) rectangle (0.6cm,0.1cm);},
        nodes near coords,
        nodes near coords align={horizontal},
        every node near coord/.append style={font=\scriptsize, /pgf/number format/fixed, /pgf/number format/precision=2},
        ]
        \addplot [fill=orange, postaction={pattern=north east lines}] coordinates {(0.3,Coffee Mate) (0.42,Pringles) (0.39,Syrup) (0.31,Detergent) (0.32,Canon Box) (0.3,Body wash) (0.3,Hot sauce) (0.41,Mom to mom) (0.35,Volumizing Shampoo) (0.3,Spray Adhesive) (0.34,All)};

        \addplot [fill=purple] coordinates {(0.58,Coffee Mate) (0.56,Pringles) (0.5,Syrup) (0.56,Detergent) (0.59,Canon Box) (0.53,Body wash) (0.49,Hot sauce) (0.51,Mom to mom) (0.55,Volumizing Shampoo) (0.47,Spray Adhesive) (0.53,All)};
      \end{axis}
    \end{tikzpicture}
  \end{adjustbox}
  \caption{\textbf{Simulation Predicted Grasp Success Probability Results:} Probability of grasp success predicted by the network for multi-fingered grasps for previously seen (left) and unseen (right) objects in simulation.\label{fig:sim_results_prob}}
  \vspace{-1mm}
\end{figure}

Figure~\ref{fig:sim_results_seen} shows the grasping success rates for our method and the baselines comparisons. We report the rate of successful grasps for each object in both the previously seen and unseen cases.
We note that our inference planner performs significantly better than the alternative approaches for both scenarios. Most importantly, for grasp initializations which would fail, the gradient-based optimization can refine these failure grasps to be successful. For example for the unseen scenario, a side grasp initialized for the ``spray adhesive'' object failed, but succeed after inference at the same object pose.
We report the average success probability predicted by the network in Figure~\ref{fig:sim_results_prob} for the initial heuristic grasps and the grasps found after inference for both the seen and unseen cases. We see that the inference is clearly improving the predicted probability, but has fairly low confidence in its prediction.



Our planner performs best with objects that can be easily enveloped from the side. It never selects precision grasps. For larger objects such as boxes the inference planner generally selects top grasps.
For shorter objects, our arm planner had a difficult time finding plans which would not collide with the table, especially for side grasps.
In terms of timing the inference-based planner takes approximately 2 to 3 seconds for each initialization, while sampling takes around 6 seconds to evaluate the 150 sampled grasp preshapes, comparable to the time to perform inference with 3 initializations. Max-eval takes 0.1 to 0.2 second for 3 initializations. The computer used to run the training and simulation experiment has an Intel-i74790k with 64GB RAM and an Nvidia GeForce GTX 970 graphics card (1.1 GHz base clock rate and 4G memory) running Ubuntu 14.04 with ROS Indigo.



\vspace{-24pt}
\subsection{Real Robot Multi-finger Grasp Inference}
\vspace{-12pt}
We evaluate our config-CNN trained on the simulated grasp data on the real robot. We compare our proposed CNN gradient ascent inference to the heuristic grasp and the ``max-eval'' grasp planner using the same heuristic initializations.
We do not compare to the regression network on the real robot, because of its poor performance in simulation.
We perform inference and max-eval following the same procedures used with the simulated data, except for one minor modification. In order to overcome the issue of planned grasps being in collision with the table, we limit the minimum \(z\)-axis value of the hand position to remain above the table.
We perform experiments on 5 objects from the YCB dataset~\cite{calli2015benchmarking}, only the ``pringles'' can objects was present in the training data; however, it was a different flavor. We evaluate each object at 5 random poses.

 \begin{figure}[ht!]
   \centering
  \begin{adjustbox}{minipage={\textwidth}}
    \begin{tikzpicture}
      \begin{axis}[
        ybar, ymin=-1, ymax=100,
        ylabel={Success Rate (\%)},
        axis lines*=left,
        symbolic x coords={Pringles, Pitcher, Lego, Mustard, Soft scrub, All},
        xtick=data,
        ticklabel style = {font=\scriptsize, rotate=45},
        legend style={font=\scriptsize, draw=none, fill=none},
        ylabel style = {font=\scriptsize},
        bar width = 6pt, height=4.5 cm, width=\linewidth,
        legend style={area legend, at={(1,1.15)}, anchor=north east, legend columns=4, },
        legend image code/.code={%
         \draw[#1] (0cm,-0.1cm) rectangle (2mm,1mm);},
        ]
        \addplot [fill=orange, postaction={pattern=north east lines}] coordinates {
          (Pringles, 43) (Pitcher, 23) (Lego,33) (Mustard,38) (Soft scrub,46)
          (All,37)};

        \addplot [fill=yellow, postaction={pattern=horizontal lines}] coordinates {
          (Pringles, 60) (Pitcher, 20) (Lego,60) (Mustard,80) (Soft scrub,60)
          (All,56)};

        \addplot [fill=purple] coordinates {
          (Pringles, 100) (Pitcher, 100) (Lego, 60) (Mustard,80) (Soft scrub,100)
          (All,88)};
        \legend{Heuristic, Max-eval, Inference}
      \end{axis}
    \end{tikzpicture}
  \end{adjustbox}
  \caption{Comparison of success rates for multi-fingered grasping on the real robot. Pringles 
    was seen in training, all other objects are previously unseen.}
   \label{fig:multifinger_inference_results}
 \end{figure}
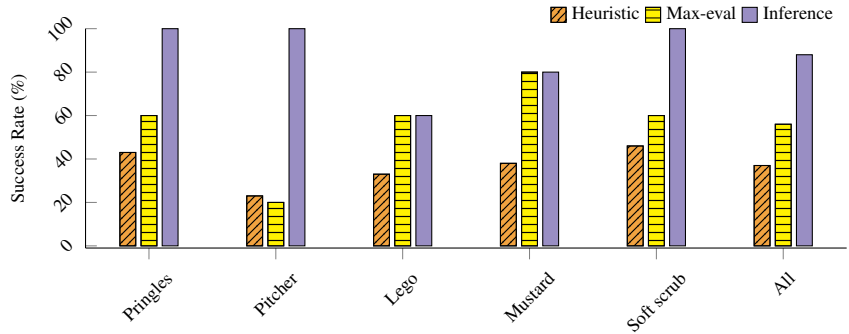

\begin{figure}[h]
   \centering
   \includegraphics[ width=0.98\linewidth]{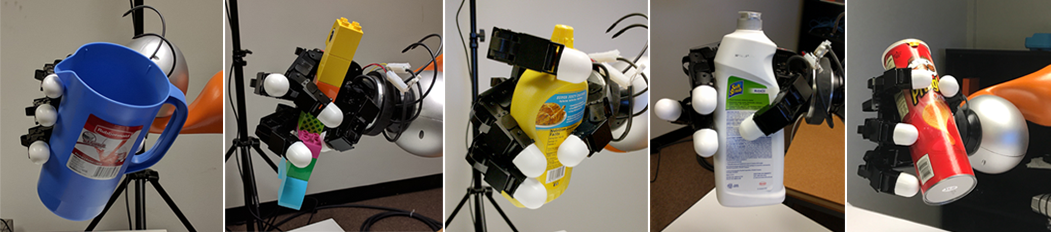}
   \caption{Examples of successful grasps generated by our inference approach to grasp planning.}
   \label{fig:inference_grasp_examples}
 \end{figure}

 Figure~\ref{fig:multifinger_inference_results} summarizes the results of these experiments.
 We note that the inference results perform better than our comparison methods in all cases. Similar with simulation, the gradient-based optimization can refine these failure grasps to be successful. For example, a side grasp initialized for the ``pitcher'' object failed, but succeed after inference at the same object pose. 
In Figure~\ref{fig:inference_grasp_examples} we show several example grasps generated by our grasp inference planner. We see that while most grasps are enveloping, different finger spreads are used to accommodate the different object geometries. The inference-based planner and max-eval have similar running time with simulation experiments run on similar hardware.

\subsection{Experiments for Two-finger Gripper Grasp Inference}
\vspace{-12pt}
As a final evaluation we apply our grasp inference approach to planning two-fingered griper grasps on the Baxter robot. We train our patches-CNN (c.f. Figure~\ref{fig:patches_cnn}) to predict two-finger gripper grasp success using the Cornell grasp dataset from~\cite{lenz2015deep}. Our patches-CNN network has similar classification performance on an offline validation dataset as the deep network results reported in~\cite{lenz2015deep}. To highlight the role of the inference procedure, and not the network, we chose to evaluate the patches-CNN, and not the config-CNN, as it is more similar to previous approaches tested on this dataset than our config-CNN.

We initialize our gradient ascent inference using four different configurations (grasp rectangle center, orientation, and gripper width) for a given object location. For all initializations we set the bounding box center to the estimated center of the segmented object and provide a random orientation. We initialize the gripper width to be 30 pixels. We perform inference initialized with all four configurations and select the configuration with highest predicted success probability to execute.
We set the maximum number of iterations for gradient ascent to 10 and set the initial step size of the backtracking line search to 0.05. We set the maximum number of iterations for the backtracking line search to 5 per ascent step and the control parameter of backtracking line search to 0.5.
We transform the rectangle found through inference to a 3D grasp pose using the same method as~\cite{lenz2015deep}. If the robot can then lift the object without dropping it to a height 0.2m, we label it successful.
\begin{figure}
  \begin{adjustbox}{valign=t,minipage={.45\textwidth}}
    \begin{tikzpicture}
      \begin{axis}[
        xbar, xmin=0, xmax=100,
        xlabel={Success Rate (\%)},
        axis lines*=left,
        symbolic y coords={Pringles, Asus Xtion, Screw Driver, Umbrella, Foam Cube, Lego Blocks, T-Shirt, Paint Can, Jello, Stapler, All},
        ytick=data,
        ticklabel style = {font=\scriptsize},
        legend style={font=\scriptsize, draw=none, fill=none},
        xlabel style = {font=\scriptsize},
        bar width = 4pt, height=6.5 cm, width=\linewidth,
        legend style={at={(1.2,-0.15)}, area legend,
          anchor=north,legend columns=-1},
        legend image code/.code={%
          \draw[#1] (0cm,-0.1cm) rectangle (0.6cm,0.1cm);},
        ]
        \addplot [fill=orange, postaction={pattern=north east lines}] coordinates {(25,Pringles) (25,Asus Xtion) (25,Screw Driver) (50,Umbrella) (75,Foam Cube) (50,Lego Blocks) (50,T-Shirt) (25,Paint Can) (25,Jello) (50,Stapler) (40,All)};

        \addplot [fill=purple] coordinates {(67,Pringles) (67,Asus Xtion) (100,Screw Driver) (67,Umbrella) (100,Foam Cube) (100,Lego Blocks) (100,T-Shirt) (67,Paint Can) (67,Jello) (100,Stapler) (84,All)};

        \legend{Initialization, Inference}
      \end{axis}
    \end{tikzpicture}
  \end{adjustbox}
  \hfill
  \begin{adjustbox}{valign=t,minipage={.45\textwidth}}
    \begin{tikzpicture}
      \begin{axis}[
        xbar, xmin=0, xmax=1,
        xlabel={Unseen Avg. Predicted Success Probability},
        axis lines*=left,
        symbolic y coords={Pringles, Asus Xtion, Screw Driver, Umbrella, Foam Cube, Lego Blocks, T-Shirt, Paint Can, Jello, Stapler, All},
        ytick=\empty,
        ticklabel style = {font=\scriptsize, /pgf/number format/fixed},
        legend style={font=\scriptsize, draw=none, fill=none},
        xlabel style = {font=\scriptsize},
        bar width = 4pt, height=6.5 cm, width=\linewidth,
        legend style={at={(0.5,1.0)}, area legend,
          anchor=south,legend columns=-1},
        legend image code/.code={%
          \draw[#1] (0cm,-0.1cm) rectangle (0.6cm,0.1cm);},
        nodes near coords,
        nodes near coords align={horizontal},
        every node near coord/.append style={font=\scriptsize, /pgf/number format/fixed, /pgf/number format/precision=2},
        ]
        \addplot [fill=orange, postaction={pattern=north east lines}] coordinates {(0.09,Pringles) (0.47,Asus Xtion) (0.24,Screw Driver) (0.62,Umbrella) (0.02,Foam Cube) (0.66,Lego Blocks) (0.49,T-Shirt) (0.01,Paint Can) (0.23,Jello) (0.05,Stapler) (0.29,All)};

        \addplot [fill=purple] coordinates {(0.97,Pringles) (0.99,Asus Xtion) (0.99,Screw Driver) (0.99,Umbrella) (0.98,Foam Cube) (0.96,Lego Blocks) (0.96,T-Shirt) (0.99,Paint Can) (0.99,Jello) (0.99,Stapler) (0.98,All)};

      \end{axis}
    \end{tikzpicture}
  \end{adjustbox}
  \caption{Success rate for executed grasps and the associated probability of grasp success predicted by the network for Baxter 2-finger grasps.\label{fig:baxter_results}}
\end{figure}
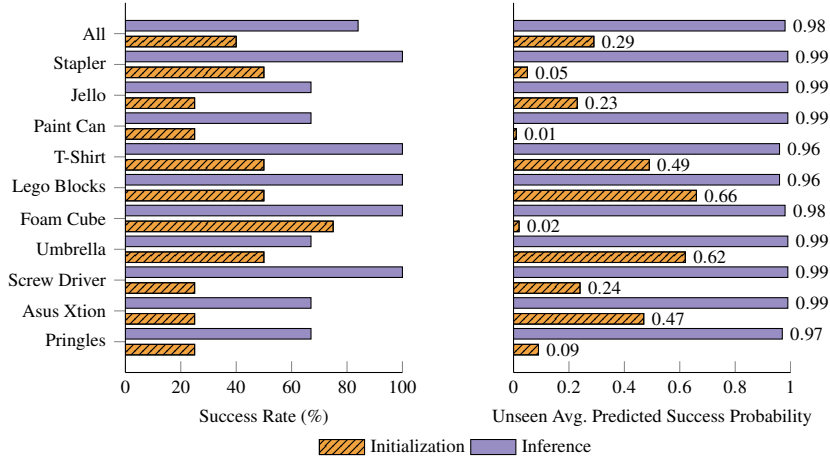
We selected 10 objects, some coming from the YCB dataset, with varying shape, texture, and visual properties comparable to the variability of the items used in~\cite{lenz2015deep}. Among these 10 objects different examples of the stapler, screw driver, and umbrella all appear in the Cornell dataset; however, the other 7 objects are entirely novel.
For each object we evaluate grasps at three random poses.
As a comparison to our inference approach we test all four random initial configurations for each object.

Figure~\ref{fig:baxter_results} shows the success rate and success probability predicted by the patches-CNN for both inference and the initial heuristic grasps for all experiments.
The inference significantly improves upon the initializations with an average success rate of $84\%$ across all objects. This performance is comparable to the success rate shown in~\cite{lenz2015deep}. However, our inference takes only 4 initializations to achieve the same success rate, far fewer than the number of grasp samples required in~\cite{lenz2015deep}. 
Inline with our multi-finger experiments, we see that the CNN gradient inference can improve initial configurations which would fail to generate successful grasps. However, the trained network is over confident in its prediction of success probability predicting at lest \(96\%\) confidence, while only being successful \(84\%\) of the time.

\vspace{-24pt}
\section{Conclusions}
\label{sec:conclusions}
We presented a novel approach for multi-fingered grasp planning formulated as probabilistic inference in a learned deep neural network.  Our planning algorithm generally achieves higher grasp success rates compared with sampling-based approaches currently used for grasping with neural networks, while still running fast enough for use in a deployed robotic system.
We showed the successful application of our grasp optimization approach on two novel neural network architectures.
Additionally, our CNN classification approach to learning grasp success allows for more data-efficient learning of grasps compared to directly predicting grasps as output of a neural network regression model.

Our multi-fingered grasping results leave room for significant improvement. Importantly any improvement to the underlying classifier will immediately be leveraged by our planner.
As immediate extensions we are examining the use of learned priors over grasp parameters, \(\theta\), as well as using priors over the neural network weights \(w\), to have better-calibrated predictions of the probability of grasp success, for reliable use within task-level planners. Finally, we wish to explore active data collection to further improve the data efficiency of our learning algorithm while also learning a wider variety of grasps.


\begin{acknowledgement}
Q.~Lu and B.~Sundaralingam were supported in part by NSF Award \#1657596.
\end{acknowledgement}
%
\vspace{-24pt}
%
%
\bibliographystyle{plainnat}
\bibliography{grasp_ref}

\end{document}